\documentclass[10pt,10pt,conference]{IEEEtran}
\IEEEoverridecommandlockouts
\usepackage{cite}
\usepackage{amsmath,amssymb,amsfonts}
\usepackage{algorithmic}
\usepackage{graphicx}
\usepackage{textcomp}
\usepackage{xcolor}
\usepackage{pifont}
\usepackage{needspace}
\usepackage{float}
\usepackage[linesnumbered,ruled,vlined]{algorithm2e}
\def\BibTeX{{\rm B\kern-.05em{\sc i\kern-.025em b}\kern-.08em
    T\kern-.1667em\lower.7ex\hbox{E}\kern-.125emX}}
\begin{document}

\title{Cog-TiPRO: Iterative Prompt Refinement with LLMs to Detect Cognitive Decline via Longitudinal Voice Assistant Commands}

\author{
    \IEEEauthorblockN{Kristin Qi\textsuperscript{1}, Youxiang Zhu\textsuperscript{1}, Caroline Summerour\textsuperscript{2}, John A. Batsis\textsuperscript{2}, Xiaohui Liang\textsuperscript{1}}
    
    \IEEEauthorblockA{\textsuperscript{1}Computer Science, University of Massachusetts, Boston, MA, USA}
    \IEEEauthorblockA{\textsuperscript{2}School of Medicine, University of North Carolina, Chapel Hill, NC, USA}
}

\maketitle

\begin{abstract}

    Early detection of cognitive decline is crucial for enabling interventions that can slow neurodegenerative disease progression. Traditional diagnostic approaches rely on labor-intensive clinical assessments, which are impractical for frequent monitoring. Our pilot study investigates voice assistant systems (VAS) as non-invasive tools for detecting cognitive decline through longitudinal analysis of speech patterns in short and unstructured voice commands. Over an 18-month period, we collected voice commands from 35 older adults, with 15 participants providing daily at-home VAS interactions. To address the challenges of analyzing these short, unstructured and noisy commands, we propose Cog-TiPRO, a framework that combines (1) LLM-driven iterative prompt refinement for linguistic feature extraction, (2) HuBERT-based acoustic feature extraction, and (3) transformer-based temporal modeling. Using iTransformer, our approach achieves 73.80\% accuracy and 72.67\% F1-score in detecting MCI, outperforming its baseline by 27.13\%. Through our LLM approach, we identify linguistic features that uniquely characterize everyday command usage patterns in individuals experiencing cognitive decline.

\end{abstract}

\section{Introduction}
Mild Cognitive Impairment (MCI), a precursor to dementia, is characterized by subtle declines in cognitive functions such as attention, memory, and language. Early detection of these changes enables timely intervention to mitigate progression to dementia. However, detection and continuous monitoring of MCI remain challenging due to the absence of gold standard biomarkers \cite{petersen2018practice}. Additionally, determining whether cognitive changes represent pathological decline requires complex clinical judgment. Currently, cognitive impairment often remains undetected until significant progression to dementia has occurred, leaving patients and families unprepared for managing its cognitive and functional impacts \cite{bradford2009missed,langa2014diagnosis}.

Traditional diagnostics for MCI rely on clinical assessments, including neuroimaging, laboratory tests, and standardized neuropsychological examinations such as the Montreal Cognitive Assessment (MoCA) \cite{hobson2015montreal}. While effective for diagnosis, these methods have limitations: They require administration by trained healthcare professionals in clinical settings, making them labor-intensive and expensive for frequent monitoring. Since speech impairments such as increased pauses, word-finding difficulty, vocal tremors, and weaker coherence are among the earliest symptom markers of cognitive decline \cite{konig2015automatic,peplinski2019objective,liang2022evaluating}, speech processing represents a promising approach for non-invasive and cost-effective MCI detection that could potentially be implemented in the everyday environment.

Prior research in cognitive monitoring through speech technology has predominantly utilized structured speech tasks (\textit{e.g.} picture descriptions) \cite{becker1994natural, luz2023multilingual,qi2025unveil}, self-reported questionnaires, and controlled protocol-based interactions \cite{becker1994pittsburgh}. Recent studies have shown potential in analyzing voice interactions with smart speakers \cite{qi2024exploiting} and phone conversations \cite{robin2020smartphone}, demonstrating associations with neuropsychological test scores. However, these approaches may not capture unstructured free-speech patterns that characterize everyday voice interactions. Furthermore, existing methods often analyze cross-sectional rather than longitudinal data, which is crucial for detecting the subtle progression of cognitive decline.

To address traditional assessment limitations, we leverage Voice Assistant Systems (VAS) such as Amazon Alexa for passive data collection and continuous monitoring. VAS devices are increasingly present in homes and provide advantages to assist older adults through voice-based, low-cost, and home-accessible features. VAS interactions capture real-world command usage that potentially provides more observational data than clinical assessments. Furthermore, VAS enables continuous data collection through both designed speech tasks and natural voice interactions during daily activities that allow for the detection of pattern changes over time.
 
Analyzing VAS interaction data for cognitive decline detection presents several challenges. VAS commands are typically short and unstructured, which are different from the spontaneous speech traditionally used in cognitive assessments. Daily VAS usage produces diverse speech samples without clinical standardization, often containing irrelevant information and noises that complicate analysis. Additionally, extracting features from everyday, at-home interactions requires identifying aspects of behavioral patterns relevant to cognitive decline while maintaining interpretability for healthcare professionals.

To address these challenges, we propose Cog-TiPRO, a framework combining: (1) large language model (LLM)-driven iterative prompt refinement for linguistic feature extraction, (2) multimodal fusion of acoustic and linguistic features, and (3) transformer-based temporal modeling. Our approach leverages LLMs to summarize linguistic features from voice commands and identify cognitive decline indicators. The iterative prompt refinement improves linguistic feature quality relevant to MCI detection. By integrating multimodal features with temporal modeling, Cog-TiPRO effectively captures the subtle progression of cognitive decline through everyday voice interactions. Our contributations can be summarized as: 
\begin{itemize}
    \item We conducted an 18-month longitudinal study collecting VAS interactions for cognitive monitoring through daily voice commands from 15 older adults in their homes.
     \item We propose Cog-TiPRO, a framework combining LLM-driven iterative prompt refinement, multimodal fusion, and time-series modeling to detect cognitive decline.
    \item Our iterative prompt optimization effectively identifies interpretable linguistic and behavioral features relevant to cognitive decline from everyday voice interactions.
    \end{itemize}
    \begin{figure*}[t]
    \centering
    \includegraphics[width=\textwidth,height=0.17\textheight]{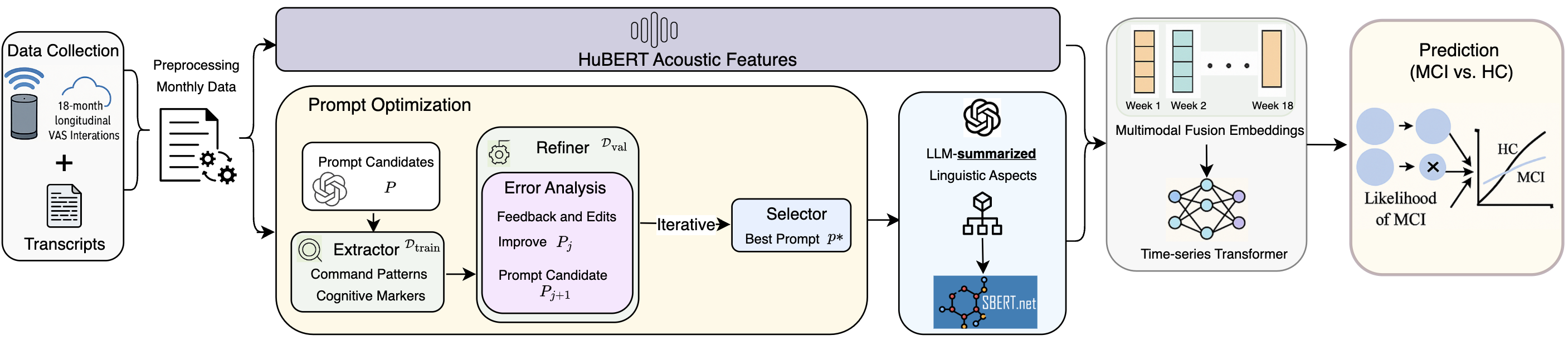}
    \vspace{2pt}
    \caption{The Cog-TiPRO framework combines (1) iterative prompt refinement optimization for linguistic feature extraction, (2) multimodal fusion of acoustic and linguistic features, and (3) transformer-based temporal modeling to detect cognitive status of MCI vs. HC.}
    \label{fig:workflow}
    \end{figure*}

\section{Related Works}
\subsection{Detecting Cognitive Decline of Dementia}
Cognitive decline detection research has primarily utilized the DementiaBank corpus for dementia detection that contains speech recordings from structured clinical interviews, such as picture description and fluency tasks \cite{becker1994pittsburgh}. These datasets present limitations: most datasets contain cross-sectional data rather than longitudinal change tracking. Traditional approaches focus on identifying paralinguistic and linguistic markers indicative of dementia, including reduced word choices, increased pauses, reduced lexical diversity, and syntactic simplification. Recent natural language processing (NLP) technologies enable automatic extraction of these features from speech transcripts, with transformer-based models such as BERT \cite{devlin2019bert} demonstrating effectiveness in capturing semantic and lexical changes associated with dementia. However, these studies apply to the onset of dementia and Alzheimer's disease rather than MCI, presenting challenges in early-stage detection.

\subsection{Large Language Model Application in Dementia Detection}
LLMs such as GPT models have transformed automatic information extraction by leveraging their extensive knowledge and in-context learning capabilities to identify linguistic patterns, even with limited training data. Recent studies by Chen \emph{et al.} \cite{chen2024profiling} and Du \emph{et al.}\cite{du2024enhancing} show that LLMs excel at filtering irrelevant information while focusing on domain-specific knowledge indicative of dementia, improving classification accuracy and interpretability in dementia detection.

\section{Voice Commands Data Collection}
\subsection{Data Collection Details}
We conducted an 18-month pilot study (2022-2024) with 35 older adults aged $\geq65$ (16 females, 19 males; 61–80 years old, mean: 72, SD: 5.1) to collect VAS voice interactions for cognitive decline detection. Participants were recruited through flyers, with informed consent obtained and procedures approved by the Institutional Review Board (IRB) to ensure ethical standards. Data collection underwent seven session-based assessments and daily in-home voice interactions. A subset of 15 participants (7 females, 8 males; age mean: 74) consented to daily in-home voice command collection. The remaining 20 participants contributed only session-based data. We utilized the 15 participants' in-home longitudinal data in further analysis.

Participants were classified as MCI or HC based on MoCA scores ($\geq$ 26 for HC, $<$ 26 for MCI). In the initial assessment, 8 were MCI and 7 were HC (labels). Ten participants had prior VAS experiences. All participants had an average MDPQ-16 (Mobile Device Proficiency Questionnaire) \cite{roque2018new} score above 4.5/5, indicating that they had adequate technical proficiency. The study combined two data collection elements:

\subsubsection{Session-Based Assessments}
Participants completed seven quarterly virtual sessions via Zoom. The initial session focused on setup and training, with subsequent sessions (1–7) including MoCA scoring, cognitive assessments, and interviews about VAS usefulness, feedback, and privacy concerns. A research assistant (RA) administered all activities.

\subsubsection{At-Home Voice Commands}
The 15 participants interacted with Alexa daily at home, with recordings and transcripts stored in Amazon's secure portal. All other data, such as consents, interviews, demographics, and MoCA scores, are stored in the HIPAA-compliant REDCap database. In the initial session, the RA provided setup training. The RA monitored participants' device usage and gave email support when participation significantly decreased. On average, each participant generated approximately 47 commands weekly.

\subsection{Data Preprocessing}
We clean the data for both audio recordings and transcripts. For audio, we remove background noise and resample all files to a 16 kHz mono channel. For transcripts, we discard error cases, including system errors (``audio could not be understood''), multiple devices in the user's home (``audio was not intended for this device''), and empty commands. To analyze command semantics, we remove the wake-up phrase ``Alexa''. One participant used three alternatives: ``computer'', ``echo'', and ``ziggy''. We identify the first words of commands, rank them by frequency, and remove the top four high-ranked words that are determined to be wake-up phrases.

\section{Preliminaries}
\subsection{Problem Formulation and Notation}
We study cognitive status assessment using an in-home longitudinal dataset containing $N = 15$ participants, each participant $i$ with a cognitive label $y^{(i)} \in \{\text{HC}, \text{MCI}\}$ determined by the clinical MoCA evaluation. Our data comprises voice interactions collected over $18$ months, indexed as $t \in \{1, 2, \ldots, 18\}$. For each participant $i$ at month $t$, we collect voice commands $A_t^{(i)}$ and their corresponding transcripts $X_t^{(i)}$, from which we derive acoustic features $\mathbf{v}_t^{(i)} \in \mathbb{R}^d$ and linguistic features $\mathbf{u}_t^{(i)} \in \mathbb{R}^e$. These are concatenated to form the multimodal representations:
\begin{equation}
\mathbf{z}_t^{(i)} = [\mathbf{v}_t^{(i)}; \mathbf{u}_t^{(i)}] \in \mathbb{R}^{d+e}
\end{equation}
The complete time-series sequence of multimodal features for participant $i$ is denoted as: 
\begin{equation}
  \mathbf{Z}^{(i)} = \{\mathbf{z}_1^{(i)}, \mathbf{z}_2^{(i)}, \ldots, \mathbf{z}_t^{(i)}\} \in \mathbb{R}^{t \times (d+e)}
\end{equation}
where each $\mathbf{z}_t^{(i)} \in \mathbb{R}^{d+e}$ is month $t$'s multimodal features.
\subsection{Cognitive Status Detection}
We formulate cognitive status detection as a sequence classification problem over the temporal progression of voice commands. Given a participant’s sequential inputs $\mathbf{Z}^{(i)}$ up to month $t$, our objective is to predict the participant’s cognitive status at $t+1$ using a time-series model:
\begin{equation}
\hat{y}^{(i)}_{t+1} = \mathcal{T}_{\Theta}(\mathbf{z}_1^{(i)}, \mathbf{z}_2^{(i)}, \ldots, \mathbf{z}_t^{(i)})
\end{equation}
where $\mathcal{T}_{\Theta}$ denotes a transformer-based time-series model with parameters $\Theta$. The core challenge is to capture patterns in command usage over time that effectively indicate MCI.

\section{Method}
In this section, we present Cog-TiPRO, which integrates three components: (1) iterative prompt refinement optimization for enhanced linguistic feature extraction, (2) multimodal fusion of acoustic and linguistic features, and (3) transformer-based time-series modeling for longitudinal analysis. Figure~\ref{fig:workflow} illustrates our proposed framework. Algorithm~\ref{alg:cogtipro} describes the full Cog-TiPRO process.
\subsection{LLM-Driven Linguistic Aspect and Feature Extraction}
To address the challenges of unstructured, short voice commands and limited training data, we develop an LLM-driven approach that identifies aspects related to cognitive markers and command usage as linguistic features. Below, we detail our iteratively optimized prompting strategy to generate feature summaries, which are then embedded using the Sentence-BERT (SBERT) all-MiniLM-L6-v2 model~\cite{reimers2019sentence}. This approach leverages the language understanding and summarization capabilities of LLMs to identify aspects relevant to cognitive decline from command usage. The iterative refinement of LLM prompts aims to produce increasingly discriminative features.
\subsubsection{Initial Prompt Design}
Our initial prompt $P_{\text{init}}$ consists of three components:
\begin{equation}
P_{\text{init}} = \{ P_{\text{context}},\; P_{\text{instruction}},\; P_{\text{exemplars}} \}
\label{eq:initial_prompt}
\end{equation}

\begin{itemize}
    \item $P_{\text{context}}$ provides background information about MCI-related cognitive markers, including lexical complexity, syntactic structure, disfluency, and semantic patterns.
    
    \item $P_{\text{instruction}}$ guides LLMs to identify relevant linguistic features by leveraging LLMs' pretrained knowledge base.
    
    \item $P_{\text{exemplars}}$ includes few-shot examples of command transcripts from both cognitive status groups (MCI vs. HC). These examples help LLMs' understanding of command patterns between the two groups.
\end{itemize}
 
\subsubsection{Iterative Prompt Optimization Details}
Our algorithm begins with an initial prompt $P_{\text{init}}$ and performs automatic prompt optimization using a minibatch of training ($\mathcal{D}_{\text{train}}$) and validation data ($\mathcal{D}_{\text{val}}$) from monthly transcripts. Using a small and diverse minibatch gives fast and low-cost feedback in short cycles that can guide effective edits without overfitting \cite{pryzant2023automatic}. The iterative refinement runs for three iterations, based on evidence that 3–5 iterations can yield effective prompts~\cite{opsahl2024optimizing}. The optimized prompt improves the extraction of linguistic features relevant to MCI detection. Specifically, the process consists of:
 
\indent\textbf{Extractor:} At each iteration $j$, the \textbf{Extractor} identifies cognitive markers and command usage patterns from minibatch $\mathcal{D}_{\text{train}}$ to generate linguistic features. We use $P_{\text{init}}$ in the first iteration, then $P_j$ (the optimized prompt) in subsequent ones to extract additional features.
\begin{equation}
\mathbf{u}_t^{(i)} = \text{LLM}_{\text{Extractor}}(X_t^{(i)}, P_j)
\end{equation}
where $X_t^{(i)}$ is the transcript for participant $i$ at month $t$.

\indent\textbf{Refiner:} The LLM refines prompts in three steps:
\begin{itemize}
    \item It uses the extracted linguistic features from the \textbf{Extractor} along with prompt $P_j$ to instruct the LLM to act as a classifier and make direct MCI vs. HC predictions, denoted as $\text{LLM}_{\text{Classifier}}$.
    \item It evaluates the LLM’s predictions using F1-score  on a validation minibatch ($\mathcal{D}_{\text{val}}$) and performs error analysis to identify specific cases where the LLM produces incorrect predictions with $P_j$.
    \item The LLM provides feedback by analyzing misclassified cases and explicitly suggesting edits to create $P_{j+1}$ that help avoid similar errors.
\end{itemize}
After these steps, the current prompt $P_j$ is updated to $P_{j+1}$ to better capture additional linguistic features:
\begin{equation}
P_{j+1} = \text{LLM}_{\text{Refiner}}\left(P_j, \text{AnalyzeErrors}(\text{errors}_j)\right)
\end{equation}
\indent\textbf{Selector:}
After three iterations, we select the best prompt $P^*$ that achieves the highest F1-score on the validation minibatch for generating linguistic feature summaries:
\begin{equation}
P^* = \arg\max_{j \in \{1,2,3\}} \mu(\mathcal{D}_{\text{val}})
\end{equation} 
where $\mu$ denotes the F1-score evaluation metric. Optimized prompt $P^*$ is used to extract linguistic features and generate summaries for these features that are then fed into SBERT.

\subsection{Multimodal Feature Embeddings}
\subsubsection{Speech Feature Embeddings} We employ the pre-trained HuBERT model~\cite{hsu2021hubert} to extract acoustic feature embeddings from monthly command audio. HuBERT is trained via self-supervision on large-scale speech corpora to capture speaker characteristics, including prosody (\textit{e.g.} pitch, rhythm, pause), spectral patterns, and articulation relevant to MCI detection. Formally, let $H(\cdot)$ be the HuBERT feature extractor. For each audio command $A_t^{(i)}$, we obtain its acoustic embedding: 
\begin{equation}
  \mathbf{v}_t^{(i)} = H(A_t^{(i)}),\ \mathbf{v}_t^{(i)} \in \mathbb{R}^{768}.
  \end{equation}
 
\subsubsection{Linguistic Feature Embeddings} 
Using the optimized prompt $P^*$, we obtain LLM-summarized linguistic features from monthly command transcripts. We use SBERT to embed these summaries as linguistic feature embeddings: 
\begin{equation}
\mathbf{u}_t^{(i)} = \text{SBERT}(\text{LLM}_{\text{Extractor}}(X_t^{(i)}, P^*)), \quad \mathbf{u}_t^{(i)} \in \mathbb{R}^{384}
\end{equation}

\subsubsection{Temporal Sequence Modeling with Transformer}
We use a time-series model to capture temporal patterns in the sequence of historical data from months 1 to $t$ ($t = 18$). The model takes as input the sequential multimodal embeddings: 
\begin{equation}
  \mathbf{Z}^{(i)} = \{\mathbf{z}_1^{(i)}, \mathbf{z}_2^{(i)}, \ldots, \mathbf{z}_t^{(i)}\} \in \mathbb{R}^{t \times 1152}
  \end{equation}
where each $\mathbf{z}_t^{(i)}$ is the multimodal features from the month $t$: $\mathbf{z}_t^{(i)} = [\mathbf{v}_t^{(i)}; \mathbf{u}_t^{(i)}]$.

We implement and compare transformer-based models such as PatchTST~\cite{nie2022time} and iTransformer~\cite{liu2023itransformer}, which have shown strong performance in modeling time-series data in several domains. The model takes $\mathbf{Z}^{(i)}$ as input and outputs a cognitive status prediction:
\begin{equation}
  \hat{y}^{(i)} = \mathcal{T}(\mathbf{Z}^{(i)}; \Theta)
  \end{equation}
where $\Theta$ denotes the learnable parameters of the transformer. 

\needspace{8\baselineskip}
The complete Cog-TiPRO can be found in Algorithm~\ref{alg:cogtipro}.
\begin{algorithm}[ht]
    \caption{Cog-TiPRO}
    \label{alg:cogtipro}
    \small
    \setlength{\algomargin}{0pt}
    \begin{algorithmic}[1]
    \REQUIRE $A_t^{(i)}$, $X_t^{(i)}$, $P_{\text{init}}$, $y^{(i)}$, MaxIter=3, $\mu$
    \STATE $P \gets P_{\text{init}}$, $P^* \gets P_{\text{init}}$, $\mu_{\text{best}} \gets 0$
    \FOR{$j = 1$ to MaxIter}
      \STATE $\mathbf{u}_t^{(i)} \gets \text{LLM}_{\text{Extractor}}(X_t^{(i)}, P)$
      \STATE $\Theta_j \gets \text{LLM}_{\text{Classifier}}(\mathbf{u}_t^{(i)}, y^{(i)} \mid \mathcal{D}_{\text{train}})$
      \STATE $\mu_j \gets \mu(\Theta_j \mid \mathcal{D}_{\text{val}})$
      \IF{$\mu_j > \mu_{\text{best}}$} \STATE $P^* \gets P$; $\mu_{\text{best}} \gets \mu_j$ \ENDIF
      \STATE $\text{errors}_j \gets \{(i,t) \mid \Theta_j(\mathbf{u}_t^{(i)}) \neq y^{(i)}\}$
      \STATE $P \gets \text{LLM}_{\text{Refiner}}(P, \text{AnalyzeErrors}(\Theta_j, \text{errors}_j))$
    \ENDFOR
    \FOR{$i,t$}
      \STATE $\mathbf{v}_t^{(i)} \gets H(A_t^{(i)})$; $\text{summary}_t^{(i)} \gets \text{LLM}_{\text{Extractor}}(X_t^{(i)}, P^*)$
      \STATE $\mathbf{u}_t^{(i)} \gets \text{SBERT}(\text{summary}_t^{(i)})$; $\mathbf{z}_t^{(i)} \gets [\mathbf{v}_t^{(i)}; \mathbf{u}_t^{(i)}]$
    \ENDFOR
    \STATE $\mathbf{Z}^{(i)} \gets \{\mathbf{z}_1^{(i)}, \ldots, \mathbf{z}_T^{(i)}\}$
    \STATE $\Theta \gets \mathcal{T}(\mathbf{Z}^{(i)}, y^{(i)})$
    \RETURN $P^*$, $\Theta$
    \end{algorithmic}
    \end{algorithm}
\section{Experimental Setup}
Given the limited sample size, we use leave-one-subject-out (LOSO) cross-validation for robust evaluation. Each experiment is run 5 times with different random seeds, and we report the average results. We evaluate performance using classification metrics: accuracy and F1-score. For prompt optimization, we use a minibatch size of 64 (51 train, 13 validation). Transformer models are trained using the AdamW optimizer with a learning rate of 1e-4, a cosine annealing schedule with 100 warm-up steps, and a batch size of 8 until reaching early stopping criteria (validation loss has no drop after 10 epochs). Maximum training is 50 epochs.

For LLMs, we obtain results by directly prompting without fine-tuning. Baseline comparisons include BERT-series models and LLMs such as Qwen2~\cite{bai2023qwen}, Flan-T5~\cite{chung2022scaling}, GLM-4~\cite{glm2024chatglm}, Llama-3~\cite{grattafiori2024llama}, and GPT-series models (3.5-turbo, 4o-mini, and 4.1-mini)~\cite{achiam2023gpt}. Training uses binary cross-entropy loss on an NVIDIA A100 GPU. All pretrained models use their default dimensional settings.

\section{Results and Discussion}
\subsection{Comparison with Baseline Models}
Table~\ref{tab:main_results} compares our Cog-TiPRO framework with various baseline models for MCI detection. Each approach is evaluated using both PatchTST and iTransformer. The results reveal the performance of different models and the impact of LLMs in identifying linguistic features.

We observe that HuBERT-based acoustic features achieve better results than many linguistic-only models. This finding suggests that HuBERT effectively extracts acoustic features that can serve as markers of cognitive function. Among non-LLM linguistic models, RoBERTa-Base achieves the strongest performance (66.67\% acc-(uracy)), 67.57\% F1), followed by BERT-Base and MedBERT (both 60.00\% acc). BioBERT shows the lowest performance (53.33\% acc), indicating that domain-specific pretraining on medical data is more beneficial than on biomedical data for cognitive assessment tasks.

For LLM-based models, we evaluate their capabilities without fine-tuning to assess their base effectiveness in identifying cognitive markers from voice commands. The results show notable performance variation. Larger models such as Qwen2-7B (66.67\% acc, 65.61\% F1) significantly outperform their smaller variants like Qwen2-1.5B (43.33\% acc, 46.38\% F1). This difference indicates that higher model capacity enables better handling of complex tasks that require contextual understanding of unstructured voice commands. However, model architecture also affects performance. Flan-T5-Base (60.00\% acc) outperforms the larger GLM-4-9B (53.33\% acc), indicating that Flan-T5’s bidirectional context modeling more efficiently captures linguistic patterns in voice command.

Llama-3-series and GPT-series models show comparable and sometimes better performance compared with other models. GPT3.5 achieves strong results (61.42\% acc, 60.63\% F1). Notably, GPT4.1-mini shows less stable results, potentially due to its optimization for different types of text generation tasks rather than analytical classification.
\setlength{\tabcolsep}{4pt}
\renewcommand{\arraystretch}{1.5} 
\begin{table}[htbp]
\caption{Comparison of Cog-TiPRO performance using PatchTST and iTransformer for accuracy (Acc) and F1-score (F1).}
\centering
\footnotesize
\begin{tabular}{|p{2.55cm}|c|c|c||c|c|}
\hline
\textbf{Methods} & \textbf{LLM} & \multicolumn{2}{c||}{\textbf{PatchTST}} & \multicolumn{2}{c|}{\textbf{iTransformer}} \\
\cline{3-6}
& & \textbf{Acc (\%)} & \textbf{F1 (\%)} & \textbf{Acc (\%)} & \textbf{F1 (\%)} \\
\hline
\multicolumn{6}{|c|}{\textit{Acoustic}} \\
\hline
HuBERT & - & \textbf{66.67} & \textbf{68.00} & 66.67 & 68.00 \\
\hline
\multicolumn{6}{|c|}{\textit{Linguistic}} \\
\hline
BERT-Base &  - & 60.00 & 63.33 & 49.99 & 53.26 \\ 
RoBERTa-Base &  - & \textbf{66.67} & 67.57 & 63.33 & 56.67   \\
MedBERT &  - & 60.00 & 63.33 & 53.33 & 60.00 \\
BioBERT &  - & 53.33 & 60.00 & 46.67 & 53.38\\
Qwen2-1.5B-Instruct & \ding{51} & 43.33 & 46.38 & 50.20 & 51.82 \\
Qwen2-7B-Instruct & \ding{51} &  \textbf{66.67}&  65.61&  63.32&  66.00\\
Flan-T5-Base & \ding{51} & 60.00 & 63.33 & 53.33 & 60.00 \\
GLM-4-9B-Chat & \ding{51} & 53.33  & 46.38 &  53.33&  51.85\\
Llama-3.2-1B & \ding{51} & 56.67 & 56.00 & 53.33 & 60.00 \\
Llama-3.1-8B-Instruct& \ding{51} & 60.55 &  50.99&  66.67& 65.61 \\
GPT3.5 & \ding{51} &  61.42& 60.63 & 60.00 & 50.60 \\
GPT4o-mini & \ding{51} &  53.93&  58.00 & 53.33 & 56.38\\
GPT4.1-mini & \ding{51} &  51.50&  46.70&  53.33&  48.83\\
\hline
\multicolumn{6}{|c|}{\textit{Multimodal}} \\
\hline
Ours (GPT3.5) & \ding{51} &  63.33& 66.20 &   70.00&  66.90\\
Ours (GPT4o-mini) & \ding{51} &   \textbf{66.67}&   63.33&  \textbf{73.80} &   \textbf{72.67}\\
Ours (GPT4.1-mini) & \ding{51} &  62.00&  61.33&  71.33& 70.67 \\
\hline
\end{tabular}
\label{tab:main_results}
\end{table}

\noindent\textbf{Effectiveness of Cog-TiPRO:  }
Cog-TiPRO consistently improves performance across most LLM variants. When combined with GPT4o-mini and iTransformer, it achieves the highest performance (73.80\% acc, 72.67\% F1), showing a 19.87\% improvement over its baseline (53.93\% acc). GPT3.5 with iTransformer ranks second (70.00\% acc, 66.90\% F1). These results indicate that advanced LLMs are more effective at identifying MCI-relevant aspects and patterns in command usage when guided by our iterative prompt refinement. This demonstrates the strength of our LLM-driven strategy with pre-trained LLM models without the need for specific fine-tuning, in contrast to previous work \cite{qi-etal-2025-umb}. Additionally, multimodal fusion outperforms single-modality baselines, suggesting that feature fusion provides complementary information to enhance feature quality and detection accuracy.

When comparing transformer architectures, iTransformer consistently outperforms PatchTST. This highlights that effective cognitive monitoring via speech requires more than just analyzing linguistic content. The integration of multimodal features and temporal patterns better captures the progression of cognitive decline. With GPT4o-mini, iTransformer achieves 73.80\% accuracy versus PatchTST's 66.67\% (7.13\% increase). iTransformer better captures temporal patterns in our data and supports its selection in our ablation study.

\subsection{Ablation Study of Main Components}
Table~\ref{tab:ablation} presents the ablation study of Cog-TiPRO components using GPT4o-mini with iTransformer. The baseline model without prompt refinement achieves the lowest performance (46.67\% acc and 43.33\% F1). This substantial performance degradation (27.13\% decrease) indicates the importance of iterative prompt optimization in generating enriched linguistic features. When prompt optimization is included, the additional incorporation of time-series transformers yields an improvement of 13.80\%. This highlights the value of incorporating sequential temporal modeling in longitudinal analysis for more accurate prediction. Although the inclusion of acoustic features shows a somewhat less significant improvement of 9.41\%, acoustic characteristics provide essential complementary information not captured by transcripts alone.
 
Combining all three components yields the best prediction performance. This finding indicates that integrating LLM-driven linguistic features, acoustic features, and temporal modeling delivers optimal identification of aspects and usage patterns in voice commands for detecting cognitive decline.
\begin{table}[htbp]
  \caption{Ablation study of Cog-TiPRO components using GPT4o-mini with iTransformer.}
  \centering
  \renewcommand{\arraystretch}{1.3} 
  \scriptsize 
  \resizebox{\columnwidth}{!}{%
  \begin{tabular}{|l|c||c|c|c|}
    \hline
    \textbf{Metric} & \textbf{Full} & \textbf{w/o Prompt} & \textbf{w/o Temporal} & \textbf{w/o Acoustic} \\
    \hline
    \rule{0pt}{1.9ex}Acc (\%) & 73.80 & 46.67 & 60.00 & 64.39 \\
    F1 (\%)  & 72.67 & 43.33 & 62.92 & 69.06 \\
    \hline
  \end{tabular}%
  }
  \label{tab:ablation}
  \end{table}
  \begin{figure*}[htbp]
    \centering
    \includegraphics[width=\textwidth, height=0.26\textheight]{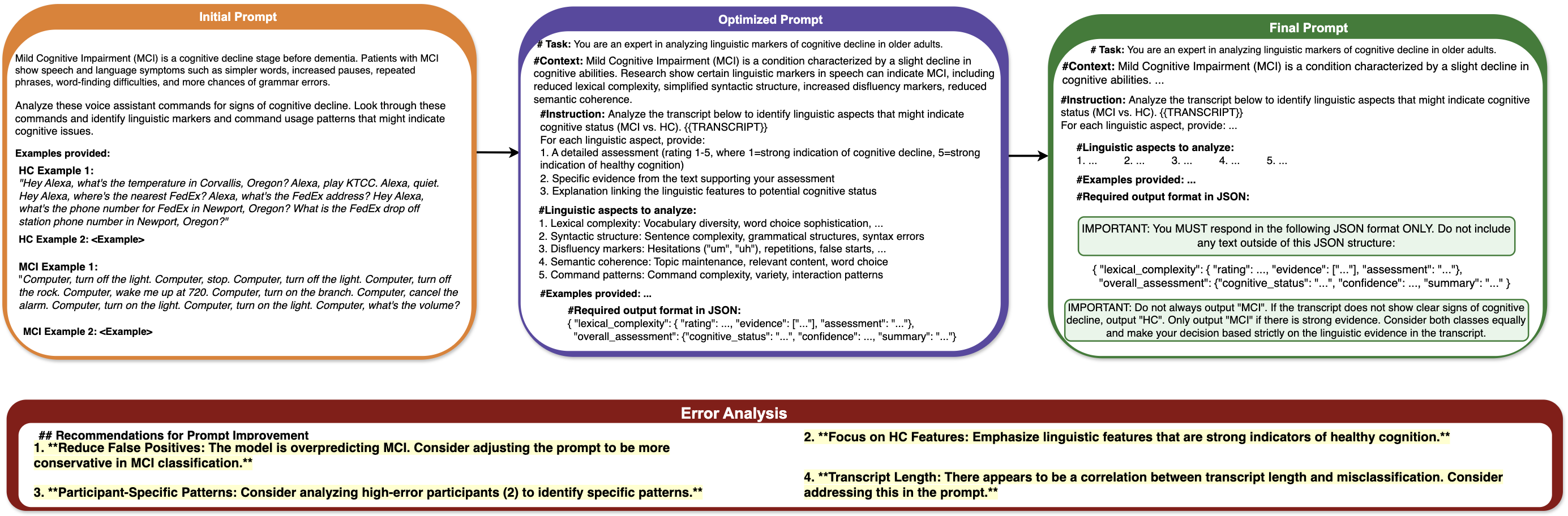}
    \vspace{1pt}
    \caption{Prompt optimization progress: initial template by a human (left), intermediate template after the first iteration (middle), and final optimized prompt after three iterations (right). The bottom panel shows an example of error analysis feedback for prompt edits.}
    \label{fig:prompt_optimization}
  \end{figure*}
  \begin{table}[htbp]
    \caption{LLM-extracted Linguistic Features Associated with MCI}
    \centering
    \renewcommand{\arraystretch}{1.3} 
    \setlength{\tabcolsep}{6pt}
    \footnotesize 
    \begin{tabular}{p{0.97\columnwidth}} 
    \hline \rule{0pt}{3ex}
    \textbf{(1) Reduced lexical diversity and word-finding difficulty:} \\
    $\bullet$ Impaired word retrieval with vague placeholders (``that thing''), re-use of identical noun phrases.\\
    $\bullet$ Commands that don't match context (``turn off the rock''), inability to access specific nouns.\\[1ex]
    \hline \rule{0pt}{3ex}
    \textbf{(2) Weak semantic / thematic coherence:} \\ 
    $\bullet$ Abrupt topic jumps ($<$10s: music $\rightarrow$ math $\rightarrow$ lights), disconnected context (``play thunderstorm'' right after ``divide 57.5 by 16.9'').\\ 
    $\bullet$ Random insertions of irrelevant words without context (``Lent''), garbled tokens, failure to maintain thematic coherence. \\[1ex]
    \hline  \rule{0pt}{3ex}
    \textbf{(3) Difficulty in self-correction:} \\ 
    $\bullet$ Starts correction but reverts to error (``wake me at 7... uh 8... no, 7''), abandoned corrections (``never mind''). \\
    $\bullet$ Persisting with incorrect commands despite system feedback, limited recognition of unsuccessful attempts and inability to request new ones.  \\[1ex]
    \hline \rule{0pt}{3ex}
    \textbf{(4) Weaker grammatical structures:} \\ 
    $\bullet$ Missing tense/number agreement, incomplete clause, omission of necessary elements, subject-verb disagreement. \\
    $\bullet$ Imperative-only style on ``play/stop/volume'' with rare ``WH''-clauses, rare conversational or informational queries. \\[0.5ex]
    \hline  \rule{0pt}{3ex}
    \textbf{(5) Repetitive command patterns and actions:} \\ 
    $\bullet$ Bursts of $\geq$3 identical requests within $<$30s, re-asking identical information despite immediate answer, no or unrelated follow-ups. \\
    $\bullet$ Rapid on/off toggles of devices, unnecessary multiple timer/alarm resets, alternating command patterns. \\[1ex]
    \hline  \rule{0pt}{3ex}
    \textbf{(6) Pragmatic function limitations:} \\ 
    $\bullet$ Day-long logs limited to alarm/light/volume, fewer exploratory queries. \\
    $\bullet$ Very short commands when the task normally elicits longer phrasing. In HC, same tasks use detailed phrasing.\\[1ex]
    \hline  \rule{0pt}{3ex}
    \textbf{(7) Disfluency overload:} \\ 
    $\bullet$ Filler pauses then command (um/uh… stop alarm), long stretches of hedging before action. \\[1ex]
    \hline
    \end{tabular}
    \label{tab:linguistic-markers}
    \end{table}
\subsection{Analysis of LLM-extracted Linguistic Features}
Table~\ref{tab:linguistic-markers} categorizes LLM-extracted features from voice commands into seven types, aligned with existing literature on cognitive decline markers. These indicators reflect distinct command usage behaviors between MCI and HC. Our analysis reveals that MCI participants exhibit difficulties increasing command variety, often repeat identical queries within short time frames, and demonstrate degradation in organizing thoughts and language. LLM-extracted command usage patterns also reveal that MCI participants are less able to respond to unsuccessful attempts, engage in information-seeking, and perform conversational follow-ups despite their awareness of device functionality. In contrast, HC participants demonstrate greater flexibility and exploratory behavior in language and topic selection, more diverse word choices, and enhanced information-seeking engagement. This comprehensive feature set presents a profile of linguistic changes of MCI that can be passively monitored through everyday voice interactions. Importantly, these early language changes are detectable in unstructured voice command settings, yet may not be captured in traditional clinical assessments.

\subsection{Visualization of Prompt Optimization and Error Analysis}
Figure~\ref{fig:prompt_optimization} illustrates our iterative prompt refinement process across three stages. The left panel shows the basic initial prompt $P_{\text{init}}$. The middle panel displays the intermediate iteration, incorporating feedback from error analysis shown at the bottom. The right panel presents the final optimized prompt $P^*$, which achieves the highest validation performance.

\section{Limitations}
Our study has several limitations. First, despite 18 months of longitudinal data, our pilot study includes a small sample size. We plan to expand the number of participants in the future work to apply Cog-TiPRO to broader and more diverse cohorts. Second, variability in home environments and reliance on voice commands may impact the quality of features related to cognitive decline and affect model performance. Our future studies will incorporate task designs and data collection that better capture established cognitive decline biomarkers to improve prediction accuracy and reliability. Furthermore, we will address participants' concerns regarding data privacy, usage, and bias in future studies. Lastly, our iterative prompt optimization increases computational overhead and inference time, which is a challenge for deployment in low-resource settings and real-time monitoring. We will focus on this problem in our future study.

\section{Conclusion}
In this paper, we demonstrate the viability of using VAS for the longitudinal data collection and analysis of voice commands to detect cognitive decline. By collecting daily interactions from 15 older adults over 18 months in their homes, we capture speech patterns that reflect cognitive functioning through our Cog-TiPRO framework. Cog-TiPRO leverages LLMs to extract linguistic features and aspects indicative of MCI from unstructured, short, and noisy voice commands. The integration of LLM-driven iterative prompt refinement, acoustic feature extraction, and temporal modeling enables more accurate detection of MCI. Using iTransformer, our approach achieves 73.80\% accuracy and 72.67\% F1-score, outperforming its baseline by 27.13\%. The linguistic features extracted not only reveal patterns specific to everyday VAS interactions but also align with established markers of MCI, such as word-finding difficulties, disfluency, and weaker coherence. This finding demonstrates the potential of passive monitoring for the early detection of cognitive decline.

\section{Acknowledgement}
This research is funded by the US National Institutes of Health National Institute on Aging R01AG067416. We thank Professor David Kotz and Professor Brian MacWhinney for providing valuable guidance.


\bibliographystyle{IEEEtran}
\bibliography{references}

\end{document}